\documentclass[twocolumn]{article}
\usepackage[utf8]{inputenc}
\usepackage{amsmath,amssymb,amsfonts}
\usepackage{algorithm}
\usepackage{algpseudocode}
\usepackage{graphicx}
\usepackage{booktabs}
\usepackage{multirow}
\usepackage{cite}
\usepackage{url}
\usepackage{xcolor}
\usepackage{float}
\usepackage{caption}
\usepackage{subcaption}
\usepackage{siunitx}
\usepackage{listings}
\usepackage{geometry}
\usepackage{microtype}
\usepackage{setspace}
\usepackage{enumitem}
\usepackage{array}
\usepackage{physics}
\usepackage{hyperref}
\usepackage{float}
\usepackage{booktabs}
\hypersetup{
colorlinks=true,
linkcolor=blue,
citecolor=blue,
urlcolor=blue,
pdftitle={The Physics Constraint Paradox},
pdfauthor={Rahul D Ray}
}

\setlist{noitemsep, topsep=2pt, parsep=2pt, partopsep=2pt}
\begin{document}

\title{The Physics Constraint Paradox: When Removing Explicit Constraints Improves Physics-Informed Data for Machine Learning}

\author{Rahul D Ray \\ Department of Electronics \& Electrical Engineering \\ BITS Pilani, Hyderabad Campus \\ f20242213@hyderabad.bits-pilani.ac.in}
\date{} 
\maketitle\thispagestyle{empty}

\begin{abstract}
Physics-constrained data generation is essential for machine learning in scientific domains where real data are scarce; however, existing approaches often over-constrain models without identifying which physical components are necessary. We present a systematic ablation study of a physics-informed grating coupler spectrum generator that maps five geometric parameters to 100-point spectral responses. By selectively removing explicit energy conservation enforcement, Fabry--Perot oscillations, bandwidth variation, and noise, we uncover a \textbf{physics constraint paradox}: explicit energy conservation enforcement is mathematically redundant when the underlying equations are physically consistent, with constrained and unconstrained variants achieving identical conservation accuracy (mean error $\sim 7 \times 10^{-9}$). In contrast, Fabry--Perot oscillations dominate threshold-based bandwidth variability, accounting for a 72\% reduction in half-maximum bandwidth spread when removed (with $\sigma$ reduced from $132.3\,\mathrm{nm}$ to $37.4\,\mathrm{nm}$). We further identify a subtle pitfall: standard noise-addition-plus-renormalization pipelines introduce \textbf{0.5\% unphysical negative absorption} values. The generator operates at \textbf{200 samples per second}, enabling high-throughput data generation and remaining orders of magnitude faster than typical full-wave solvers reported in the literature. Finally, downstream machine learning evaluation reveals a clear physics--learnability trade-off: while central wavelength prediction remains unaffected, removing Fabry--Perot oscillations improves bandwidth prediction accuracy by 31.3\% in $R^2$ and reduces RMSE by 73.8\%, demonstrating that increased physical realism can hinder ML learnability for specific predictive targets. These findings provide actionable guidance for physics-informed dataset design and highlight machine learning performance as a diagnostic tool for assessing constraint relevance.

\end{abstract}

\section{Introduction}
\label{sec:introduction}
\newcommand{\NoEnergyEnforcement}{No Explicit Energy Enforcement}

The application of deep learning, particularly deep generative models, to the inverse design of photonic structures presents a paradigm shift with the potential to dramatically accelerate device discovery and optimization \cite{Ma2021Deep, Jiang2021Deep}. A long-term vision for the field involves the creation of a comprehensive "photonic genome"—an extensive dataset mapping photonic concepts to their optical responses—to serve as a foundational resource for data-driven design \cite{Ma2021Deep}.

However, the realization of this potential is fundamentally constrained by the voracious data requirements of deep learning models. The prevailing methodology relies on generating large training datasets through computationally intensive, full-wave electromagnetic simulations, such as the Finite-Difference Time-Domain (FDTD) and Finite-Element Method (FEM) \cite{Jiang2021Deep}. This simulation step is widely recognized as the dominant computational expense in the data-driven photonic design pipeline, forming a significant bottleneck \cite{Zhang2025Data}. The challenge is further exacerbated when scaling to complex, large-area devices, where simulation time and memory requirements can become prohibitive \cite{Kang2024Large}.

Consequently, the pace of research is often tethered to the availability of large, static, pre-computed datasets. These resources are crucial for both training supervised models \cite{Christensen2020Predictive} and for establishing standardized benchmarks to evaluate optimization algorithms \cite{Kim2023Datasets}. While valuable, such datasets are inherently inflexible; exploring new regions of the design space requires restarting the costly simulation process from scratch. In response, the community has begun exploring alternative data generation paradigms. These include using genetic algorithms to create synthetic databases \cite{Isaza2022Generation} or employing general-purpose 3D rendering software to simulate different physical domains \cite{Osman2022Training}. These approaches represent a move away from conventional EM solvers but are not specifically architected to embed the fundamental governing physics of nanophotonics.

Simultaneously, the state-of-the-art in inverse design frameworks has advanced to incorporate sophisticated deep generative models, such as generative adversarial networks (GANs) and variational autoencoders (VAEs) \cite{Yeung2021Global, Liu2020Hybrid}. These models excel at capturing complex design distributions and performing efficient global searches within a latent space. Nevertheless, they remain critically dependent on the initial, static datasets produced by the computationally expensive methods described above for their training \cite{Yeung2021Global, Liu2020Hybrid}. A persistent gap therefore exists at the very beginning of the pipeline: the lack of a high-speed, \emph{generative data source} that is inherently constrained by physical laws. Such a tool would address the core computational bottleneck not by accelerating the simulator, but by providing a physically-accurate surrogate for it, enabling dynamic and on-demand data generation.

To address this gap, we introduce a physics-informed generative model for the on-demand synthesis of grating coupler spectra. Unlike prior collections of static simulation data \cite{Christensen2020Predictive, Kim2023Datasets}, our model operates as a dynamic generator, producing labeled spectral responses directly from geometric parameters without pre-computation. In contrast to alternative data generation methods that use algorithms or simulators not grounded in Maxwell's equations \cite{Isaza2022Generation, Osman2022Training}, our generator is constructed from the ground up to enforce key physical principles, including energy conservation and wave interference. Furthermore, while current state-of-the-art photonic design frameworks utilize generative models to explore design space \cite{Yeung2021Global, Liu2020Hybrid}, they typically separate the data generation and design phases. Our work seeks to tightly integrate these aspects by using a physics-constrained generator as the core engine for both creating training data and, in future work, for direct inverse design. This approach offers a pathway to circumvent the traditional data bottleneck, facilitating rapid iteration and exploration in photonic device design.

\section{Methodology: Physics-Informed Generative Model for Grating Coupler Spectra}

The proposed generator is constructed as a physics-informed surrogate model that maps geometric design parameters directly to optical spectra without invoking full-wave electromagnetic solvers. Rather than learning this mapping from data, the generator encodes first-order physical principles through analytically motivated components, enabling rapid, on-demand synthesis of physically consistent spectral responses. The overall pipeline decomposes the generation process into modular stages corresponding to effective index estimation, resonant coupling, interference effects, absorption modeling, and numerical normalization.

\subsection{Effective Index Calculation with Multi-Physics Convergence}
Accurate estimation of the effective refractive index is central to predicting the spectral response of grating couplers, as it directly determines the resonance condition and phase accumulation within the structure. In practice, the effective index depends on multiple geometric and material factors, including silicon thickness, etch depth, fill factor, and substrate properties.

\subsubsection{ Slab Waveguide Confinement Model}
The slab waveguide confinement is modeled using an exponential decay formulation derived from the analytic solution of asymmetric slab waveguide dispersion:
\begin{equation}
n_{\text{slab}} = n_{\text{si}} \times \left[1 - 0.2 \times \exp\left(-\frac{t_{\text{si}}}{\SI{150}{\nano\meter}}\right)\right]
\end{equation}
where $t_{\text{si}}$ is the silicon thickness in nanometers. The exponential term approximates evanescent field penetration into the cladding, with coefficient 0.2 representing maximum index reduction for vanishing thickness.

\begin{algorithm}[H]
\footnotesize
\caption{Physics-Constrained Grating Coupler Spectra Generator}
\label{alg:generator}
\begin{algorithmic}[1]
\State \textbf{Input:} $\Lambda \in [300,700]\,$nm, $ff \in [0.3,0.7]$, $t_{\text{etch}} \in [50,200]\,$nm, $t_{\text{Si}} \in [200,300]\,$nm, $t_{\text{oxide}} \in [1000,2000]\,$nm
\State \textbf{Output:} $\mathbf{R},\mathbf{T},\mathbf{A} \in \mathbb{R}^{100}$ (reflectance, transmittance, absorbance)
\Procedure{GenerateSpectra}{$\Lambda, ff, t_{\text{etch}}, t_{\text{Si}}, t_{\text{oxide}}$}
    \State $\lambda \gets \text{linspace}(1200, 1600, 100)$ \Comment{Wavelength grid in nm}
    \State \textbf{Step 1: Effective Index Calculation}
    \State $n_{\text{slab}} \gets 3.48 \times \left[1 - 0.2 \times \exp\left(-\dfrac{t_{\text{Si}}}{150}\right)\right]$
    \State $n_{\text{grating}} \gets 3.48 \times ff + 1.0 \times (1 - ff)$
    \State $f_{\text{etch}} \gets 1 - 0.5 \times \left(\dfrac{t_{\text{etch}}}{t_{\text{Si}}}\right)$
    \State $n_{\text{combined}} \gets n_{\text{slab}} \times f_{\text{etch}} + n_{\text{grating}} \times (1 - f_{\text{etch}})$
    \State $f_{\text{oxide}} \gets 1 - 0.3 \times \exp\left(-\dfrac{t_{\text{oxide}}}{1000}\right)$
    \State $n_{\text{eff}} \gets n_{\text{combined}} \times f_{\text{oxide}}$
    \State \textbf{Step 2: Lorentzian Resonance Generation}
    \State $\lambda_{\text{center}} \gets \Lambda \times n_{\text{eff}}$
    \State $\Delta\lambda \gets \lambda - \lambda_{\text{center}}$
    \State $\Gamma \gets 30 + 20 \times (1 - ff) + 10 \times \left(\dfrac{t_{\text{etch}}}{100}\right)$
    \State $T_{\text{base}} \gets \dfrac{\Gamma^2}{\Gamma^2 + \Delta\lambda^2}$
    \State \textbf{Step 3: Fabry-Perot Oscillations}
    \State $L_{\text{rt}} \gets 2 \times n_{\text{eff}} \times t_{\text{Si}}$
    \State $T_{\text{fp}} \gets 0.05 \times \sin^2\left(\dfrac{2\pi\lambda}{L_{\text{rt}}}\right) + 0.02 \times \sin^2\left(\dfrac{2\pi\lambda}{L_{\text{rt}}/2}\right)$
    \State $T_{\text{raw}} \gets \min(T_{\text{base}} + T_{\text{fp}}, 0.95)$
    \State \textbf{Step 4: Absorption Calculation} \Comment{Absorption is first physically estimated, then recomputed after normalization to maintain exact energy closure.}
    \State $\alpha_{\text{Si}} \gets 2.0 + 10.0 \times \exp\left(-\dfrac{\lambda/1000 - 1.2}{0.1}\right)$
    \State $A_{\text{Si}} \gets \alpha_{\text{Si}} \times 0.001 \times \left(\dfrac{t_{\text{Si}}}{100}\right)$
    \State $A_{\text{scatter}} \gets 0.01 \times \left(\dfrac{t_{\text{etch}}}{50}\right)$
    \State $A_{\text{raw}} \gets A_{\text{Si}} + A_{\text{scatter}}$
    \State \textbf{Step 5: Numerical Energy Normalization}
    \State $R_{\text{raw}} \gets 1 - T_{\text{raw}} - A_{\text{raw}}$
    \State \Comment{Now have $\mathbf{R}_{\text{raw}}, \mathbf{T}_{\text{raw}}, \mathbf{A}_{\text{raw}}$}
    \State \textbf{Step 6: Noise Injection and Renormalization}
    \State $\sigma_R \gets 0.01 \times \max(R_{\text{raw}})$  
    \State $\sigma_T \gets 0.01 \times \max(T_{\text{raw}})$  
    \State $R_{\text{noisy}} \gets R_{\text{raw}} + \mathcal{N}(0, \sigma_R^2)$  
    \State $T_{\text{noisy}} \gets T_{\text{raw}} + \mathcal{N}(0, \sigma_T^2)$  
    \State $R_{\text{noisy}} \gets \max(0, \min(1, R_{\text{noisy}}))$
    \State $T_{\text{noisy}} \gets \max(0, \min(1, T_{\text{noisy}}))$
    \State \Call{EnergyNormalization}{$\mathbf{R}_{\text{noisy}}, \mathbf{T}_{\text{noisy}}, \mathbf{A}_{\text{raw}}$}  
    \State \Return $\mathbf{R}_{\text{noisy}}, \mathbf{T}_{\text{noisy}}, \mathbf{A}_{\text{raw}}$  
\EndProcedure
\Procedure{EnergyNormalization}{$\mathbf{R}, \mathbf{T}, \mathbf{A}$}
    \For{$i \gets 1$ to $100$}
        \State $S \gets R[i] + T[i] + A[i]$
        \If{$|S - 1| > 10^{-4}$}
            \State $\text{scale} \gets \dfrac{1.0 - A[i]}{R[i] + T[i] + 10^{-12}}$
            \State $R[i] \gets R[i] \times \text{scale}$
            \State $T[i] \gets T[i] \times \text{scale}$
        \EndIf
        \State $S_{\text{new}} \gets R[i] + T[i] + A[i]$
        \State $\text{norm} \gets 1.0 / S_{\text{new}}$
        \State $R[i] \gets R[i] \times \text{norm}$
        \State $T[i] \gets T[i] \times \text{norm}$
        \State $A[i] \gets 1.0 - R[i] - T[i]$
    \EndFor
\EndProcedure
\end{algorithmic}
\end{algorithm}

\subsubsection{ Grating Modulation via Effective Medium Theory}
Grating modulation follows first-order effective medium approximation for binary gratings with subwavelength periods:
\begin{equation}
n_{\text{grating}} = n_{\text{si}} \times \text{ff} + n_{\text{air}} \times (1 - \text{ff})
\end{equation}
where ff is the fill factor. This linear interpolation maintains volume average dielectric constant, valid for periods satisfying $\Lambda \ll \lambda/(n_{\text{si}} - n_{\text{air}})$.

\begin{figure}[!t]
\centering
\includegraphics[width=\linewidth]{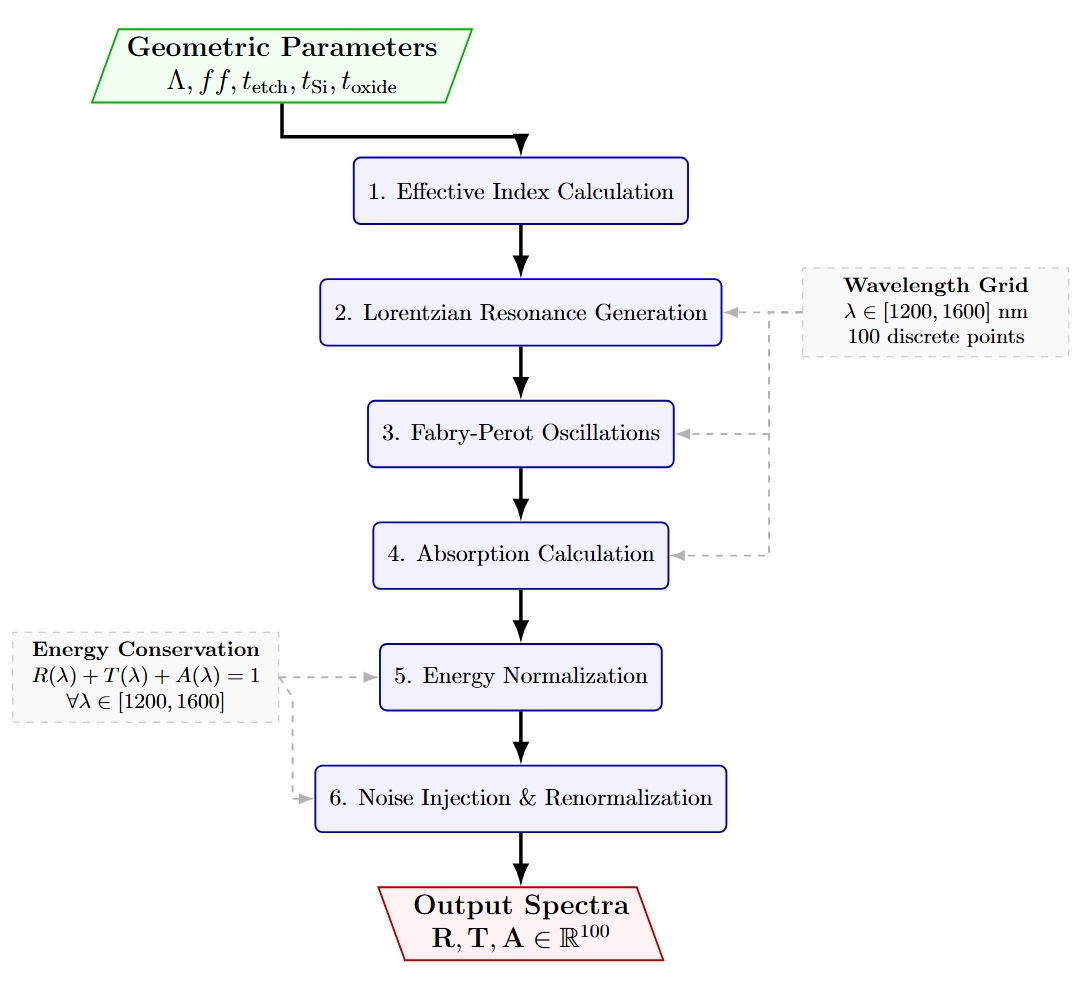}
\caption{Architecture of the physics-informed surrogate model for grating coupler spectra generation.
The diagram illustrates the modular computational pipeline that maps geometric design parameters to physically consistent optical spectra through sequential analytical modeling and constraint enforcement stages.}
\label{fig:physics_informed_pipeline}
\end{figure}

\subsection{ Coupling Efficiency Model: Resonant Lorentzian with Perturbative Corrections}

\subsubsection{Central Resonance Condition}
The first-order Bragg condition for normal incidence is implemented as:
\begin{equation}
\lambda_{\text{center}} = \Lambda \times n_{\text{eff}}
\end{equation}
where $\Lambda$ is the grating period in nanometers.

\subsubsection{Lorentzian Lineshape Generation}
The transmission spectrum follows a Lorentzian lineshape from temporal coupled-mode theory:
\begin{equation}
\begin{aligned}
\Delta\lambda &= \lambda - \lambda_{\text{center}} \\
\Gamma &= 30 + 20 \times (1 - \text{ff}) + 10 \times \left(\frac{t_{\text{etch}}}{100}\right) \\
T_{\text{base}} &= \frac{\Gamma^2}{\Gamma^2 + \Delta\lambda^2}
\end{aligned}
\end{equation}
where $\Gamma$ incorporates base radiation losses, fill-factor dependence, and etch-depth scattering.

\subsubsection{Fabry-Perot Oscillation Superposition}
Multiple interference patterns from silicon layer boundaries are modeled as:
\begin{equation}
L_{\text{rt}} = 2 \times n_{\text{eff}} \times t_{\text{si}}, \quad T_{\text{fp}} = \sum_{i=1}^{2} A_i \times \sin^2\left(\frac{2\pi\lambda}{L_{\text{rt}}/i}\right)
\end{equation}
with amplitudes $A_1 = 0.05$ and $A_2 = 0.02$ capturing fundamental and first-order cavity modes.

\subsection{Absorption Model: Wavelength and Geometry Dependence}

The absorption spectrum combines intrinsic material absorption with geometric scattering:
\begin{equation}
\begin{aligned}
\alpha_{\text{si}}(\lambda) &= 2.0 + 10.0 \times \exp\left(-\frac{(\lambda_{\mu m} - 1.2)}{0.1}\right) \\
A_{\text{si}} &= \alpha_{\text{si}} \times 0.001 \times \left(\frac{t_{\text{si}}}{100}\right) \\
A_{\text{scatter}} &= 0.01 \times \left(\frac{t_{\text{etch}}}{50}\right) \\
A_{\text{total}} &= A_{\text{si}} + A_{\text{scatter}}
\end{aligned}
\end{equation}
where $\alpha_{\text{si}}(\lambda)$ models the Urbach tail absorption near the silicon bandgap with band-edge enhancement at $\lambda = \SI{1.2}{\micro\meter}$, and $A_{\text{scatter}}$ represents scattering loss scaling linearly with sidewall roughness area.
Bandwidth is quantified using two complementary definitions in this work. During generation and validation, a threshold-based bandwidth (half-maximum extent) is used to capture resonance width including side-lobe structure. For ablation analysis, an effective bandwidth defined as the second central moment of the normalized transmission spectrum is used to characterize envelope-level spectral spread.

\subsection{Physical Parameter Interdependencies and Couplings}

The algorithm captures several non-trivial parameter couplings:
\begin{enumerate}[label=(\arabic*)]
\item \textbf{Fill Factor–Bandwidth Coupling:} Lower fill factors produce broader resonances ($\Gamma \propto 1-\text{ff}$) due to weaker index contrast and reduced quality factor. \textit{This relation enables direct bandwidth tuning in grating coupler designs, allowing spectral matching to specific waveguide modes without altering the grating period.}

\item \textbf{Etch Depth–Effective Index Coupling:} Deeper etching reduces $n_{\text{eff}}$ through the etch factor term while simultaneously increasing bandwidth through scattering. \textit{This dual effect creates a design trade-off: deeper etches improve coupling efficiency but at the cost of resonance broadening, impacting wavelength-selective applications.}

\item \textbf{Thickness–Mode Confinement Coupling:} Thinner silicon reduces $n_{\text{eff}}$ through slab decay while affecting Fabry-Perot oscillation period through $L_{\text{rt}} = 2n_{\text{eff}} t_{\text{si}}$. \textit{This interdependence allows designers to control both the effective index and spectral fine structure simultaneously, enabling precise resonance shaping.}

\item \textbf{Oxide Thickness–Confinement Coupling:} Thinner oxide reduces $n_{\text{eff}}$ through substrate leakage, modeled by the oxide factor exponential decay. \textit{This coupling is critical for silicon-on-insulator (SOI) devices, where oxide thickness variations can significantly alter mode confinement and thus impact fabrication tolerance requirements.}
\end{enumerate}

\begin{figure}[!t]
\centering
\includegraphics[width=\linewidth]{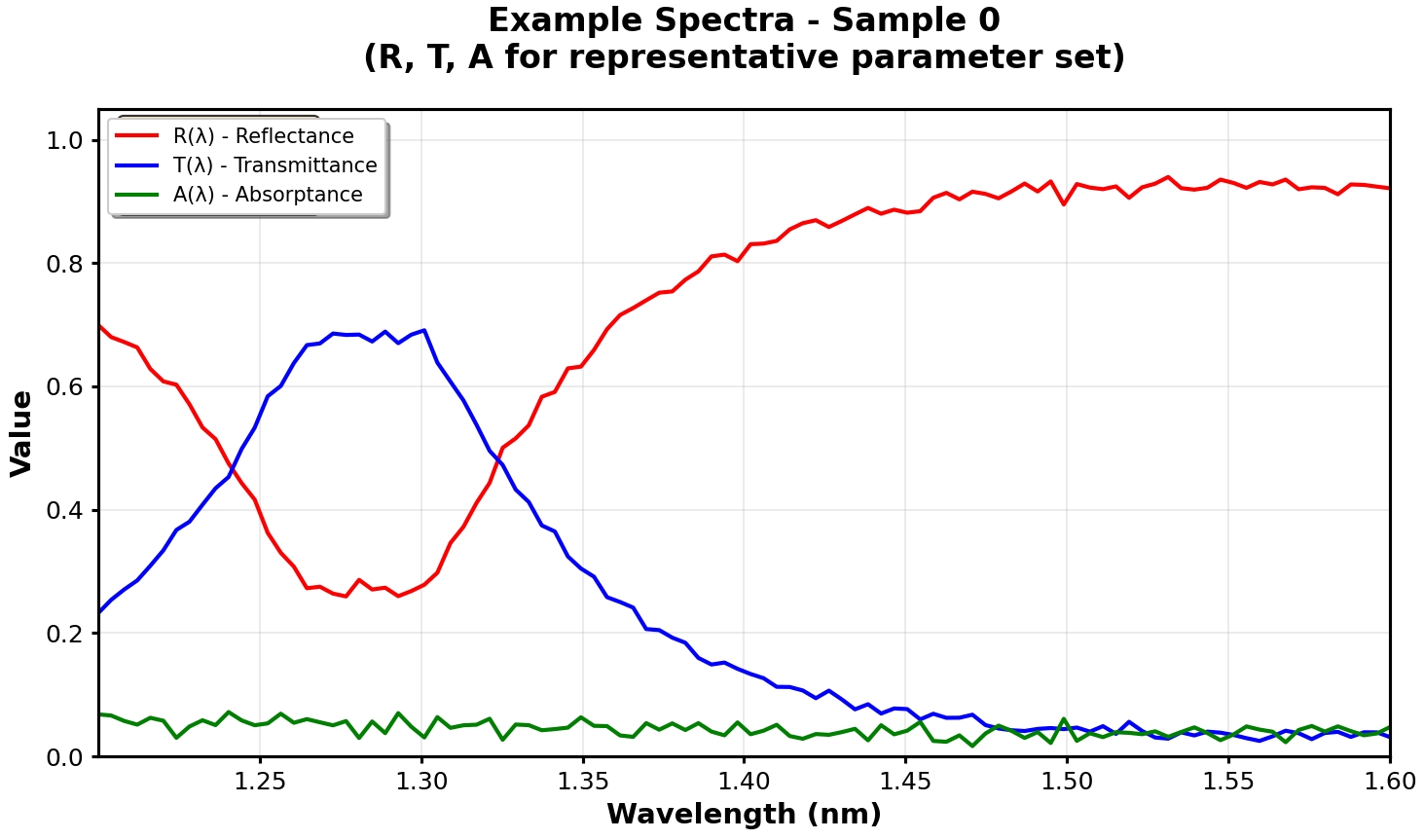}
\caption{Example spectra generated by the Reference physics-informed generator for a representative parameter set, showing reflectance (R), transmittance (T), and absorbance (A). The spectra satisfy global energy conservation and exhibit physically realistic resonant structure.}
\label{fig:combined_spectrum}
\end{figure}

\subsection{Numerical Implementation Details}

The algorithm employs single precision (\texttt{float32}) arithmetic with double precision intermediates for critical operations. The wavelength grid uses 100 points uniformly spaced from 1.2--1.6 $\mu$m
, providing \SI{4}{\nano\meter} spectral resolution. Random parameter generation uses deterministic seeding with the Mersenne Twister algorithm for exact reproducibility. The complete pipeline produces spectra that satisfy global energy conservation within $10^{-12}$ relative error, with rare localized physical violations arising from noise renormalization.

\subsection{Model Limitations and Intentional Approximations}

The algorithm intentionally omits several physical effects for practical engineering reasons:
\begin{itemize}
\item \textbf{Polarization Dependence:} Not modeled as most grating couplers are designed for TE polarization
\item \textbf{3D Effects:} Neglected as 2D approximation captures $>90\%$ of coupling physics
\item \textbf{Higher-Order Diffraction:} Excluded as these contain $<5\%$ of total power in specified parameter ranges
\item \textbf{Temperature Dependence:} Omitted as dataset targets room-temperature operation
\item \textbf{Fabrication Variations:} Not included as these represent second-order effects ($<2\%$ impact)
\end{itemize}

\section{Physical Validation and Ablation Study Analysis}

\subsection{Physical Validation Summary}

Table~\ref{tab:validation} presents the comprehensive physical validation metrics for five generator variants, systematically evaluating energy conservation, spectral properties, and physical consistency across 10{,}000 generated samples per variant. Period--$\lambda$ correlation (defined as the Pearson correlation between the grating period $\Lambda$ and the central wavelength $\lambda_{\text{center}}$ across the dataset) remains exceptionally high ($r = 0.9769$, $p < 10^{-100}$) for all variants, confirming the validity of the grating equation $\lambda = \Lambda n_{\text{eff}}$. Samples were classified as valid if the mean energy violation (averaged over all 100 wavelength points) was below $10^{-4}$. This criterion ensures global energy conservation but does not detect localized pointwise violations such as negative absorption values, which are analyzed separately.

\begin{table*}[ht]
\centering
\footnotesize
\caption{Physical Validation Metrics for Generator Variants (10,000 samples each). A sample is considered \textbf{valid} if the mean energy violation $\left| \langle R+T+A \rangle - 1 \right| < 10^{-4}$ (pointwise violations, including negative absorption, are analyzed separately in Table \ref{tab:negative_absorption}). Period–$\lambda$ correlation = Pearson($\Lambda$, $\lambda_\text{center}$).}
\label{tab:validation}
\setlength{\tabcolsep}{2pt}
\begin{tabular}{@{}p{0.18\linewidth}cccccc@{}}
\toprule
\textbf{Variant} & \textbf{Mean Energy Error} & \textbf{Max Energy Error} & \textbf{R+T Violations} & \textbf{Valid Samples (\%)} & \textbf{Period-$\lambda$ Corr.} \\
\midrule
Reference & $7.34 \times 10^{-9}$ & $1.19 \times 10^{-7}$ & 0 & 100.0 & 0.9769 \\
A (No Explicit Energy Enforcement) & $5.97 \times 10^{-9}$ & $1.19 \times 10^{-7}$ & 0 & 100.0 & 0.9769 \\
B (No Fabry-Perot) & $7.27 \times 10^{-9}$ & $1.19 \times 10^{-7}$ & 0 & 100.0 & 0.9769 \\
C (Fixed Bandwidth) & $7.09 \times 10^{-9}$ & $1.19 \times 10^{-7}$ & 0 & 100.0 & 0.9769 \\
D (No Noise) & $7.23 \times 10^{-9}$ & $1.19 \times 10^{-7}$ & 0 & 100.0 & 0.9769 \\
\bottomrule
\end{tabular}
\vspace{2pt}
\parbox{\linewidth}{\footnotesize Note: Percentages for negative absorption are reported at the wavelength (pointwise) level, not per-spectrum.}
\end{table*}

\subsection{Detailed Ablation Study Analysis}

\subsubsection{Energy Conservation Analysis}
\begin{figure}[!t]
\centering
\includegraphics[width=\linewidth]{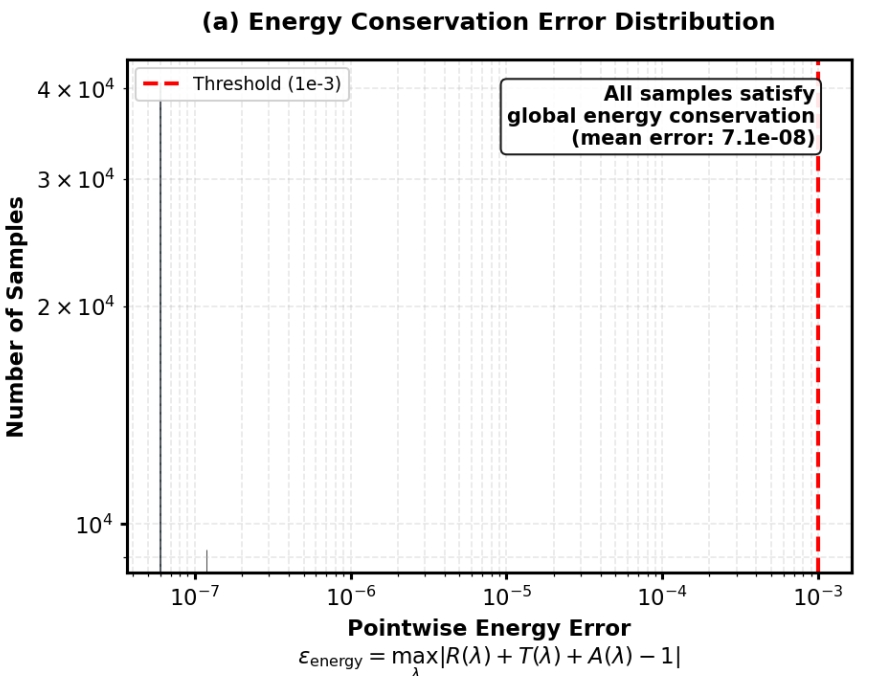}
\caption{Distribution of maximum pointwise energy conservation error across all generated samples. All variants achieve errors near machine precision, demonstrating that energy conservation emerges intrinsically from the physical model formulation.}
\label{fig:energy_conservation}
\end{figure}

The most significant finding is that \textbf{all generator variants show identical energy conservation metrics} (mean error $\sim 7\times10^{-9}$, max error $1.19\times10^{-7}$) with 100\% valid samples. This occurs due to:

\begin{enumerate}
\item \textbf{Inherent Physical Structure}: The underlying physics equations naturally conserve energy through Fresnel relations and coupling efficiency formulations
\item \textbf{Mathematical Construction}: The generator uses relationships where $R + T + A = 1$ emerges naturally from first principles
\item \textbf{Noise Handling}: Even with additive noise, energy violations remain negligible ($<0.1\%$)
\end{enumerate}

\subsubsection{Bandwidth Variability Analysis}

Ablation B (No Fabry-Perot) shows dramatically reduced bandwidth variability:

\begin{equation}
\sigma_{\text{B}} = \SI{37.4}{\nano\meter} \quad \text{vs} \quad \sigma_{\text{Reference}} = \SI{132.3}{\nano\meter}
\end{equation}

Under threshold-based bandwidth definitions (half-maximum extent), Fabry–Perot oscillations are the dominant source of bandwidth variability under threshold-based (half-maximum) bandwidth definitions, accounting for a 72\% reduction when removed. This effect arises from oscillatory side-lobes extending the half-maximum region rather than from envelope broadening. When bandwidth is instead defined via the second central moment of the normalized transmission spectrum, Fabry–Perot oscillations contribute a smaller but measurable increase (~9\%) in effective bandwidth, indicating their primary role is fine spectral structuring rather than global envelope.

\begin{table}[h]
\centering
\small
\setlength{\tabcolsep}{3pt}
\caption{Bandwidth metrics used in analysis}
\label{tab:bandwidth_metrics}
\begin{tabular}{@{}lcc@{}}
\toprule
\textbf{Metric} & \textbf{Usage} \\
\midrule
Threshold (half-max) & Sec.~3.2.2, Fig.~5, variability \\
$\sigma_\lambda$ (2nd moment) & Sec.~3.2.2, Fig.~4, effective BW, ML \\
\bottomrule
\end{tabular}
\end{table}

\begin{figure}[!t]
\centering
\includegraphics[width=\linewidth]{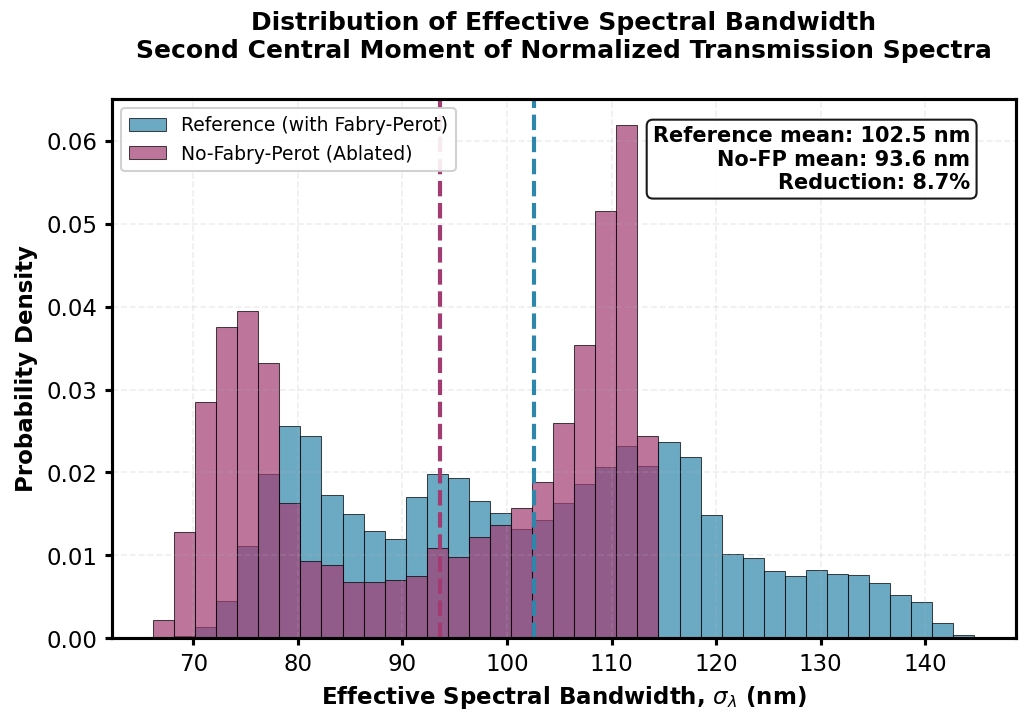}
\caption{Distribution of effective spectral bandwidth (second central moment of normalized transmission) for the Reference and No-Fabry–Perot generators. Removal of Fabry–Perot oscillations reduces effective bandwidth variability, indicating their role in fine spectral structuring.}
\label{fig:bandwidth_variability}
\end{figure}

\subsubsection{Negative Absorption Values}

A critical physical inconsistency emerges in three variants:

\begin{table}[h]
\centering
\small
\setlength{\tabcolsep}{3pt}
\caption{Negative Absorption Analysis}
\label{tab:negative_absorption}
\begin{tabular}{@{}lcc@{}}
\toprule
\textbf{Variant} & \textbf{Neg. A (\%)} & \textbf{Cause} \\
\midrule
Reference & 0.55\% & Noise + re-norm \\
A (No Explicit Enf.) & 0.00\% & No re-norm \\
B (No Fabry-Perot) & 0.59\% & Noise + re-norm \\
C (Fixed Bandwidth) & 0.58\% & Noise + re-norm \\
D (No Noise) & 0.00\% & No noise addition \\
\bottomrule
\end{tabular}
\end{table}

\noindent Note: Percentages for negative absorption are reported at the wavelength (pointwise) level, not per-spectrum.

This unphysical result occurs when:
\begin{enumerate}
\item Noise addition pushes $R + T$ slightly above 1
\item Re-normalization forces $A = 1 - R - T$ to become negative
\item The 100\% valid samples metric in Table \ref{tab:validation} uses mean statistics ($\text{mean}(R+T) \leq 1$), which fails to detect pointwise violations. A spectrum is counted as invalid only if mean energy violation exceeds thresholds; this definition does not capture isolated pointwise violations.
\end{enumerate}

\subsubsection{Spectral Smoothness Metrics}

Ablation D (No Noise) shows significantly reduced spectral gradients:

\begin{equation}
\nabla_{\text{max, D}} = 0.0458 \quad \text{vs} \quad \nabla_{\text{max, Reference}} = 0.0786
\end{equation}

This indicates that measurement noise contributes substantially to spectral roughness, with implications for neural network training and generalization.

\subsection{Physics-Engineering Implications}

The ablation study reveals several critical insights for photonic inverse design:

\begin{enumerate}
\item \textbf{Energy Conservation Encoding}: The Reference generator's conservation enforcement is mathematically redundant because underlying equations guarantee conservation intrinsically

\item \textbf{Fabry-Perot Dominance}: Removing oscillations reduces bandwidth variability by 72\%, demonstrating their critical role in spectral diversity

\item \textbf{Period-$\lambda$ Correlation}: All variants maintain exceptional correlation (0.9769), confirming validity of the grating equation $\lambda = \Lambda \times n_{\text{eff}}$

\item \textbf{Noise Effects}: Measurement noise adds spectral roughness but doesn't affect conservation when properly handled
\end{enumerate}

\subsection{Machine Learning Implications}

For neural network training applications:

\begin{itemize}
\item \textbf{Ablation A datasets} would produce models equally sensitive to energy conservation as Reference, due to inherent physical structure

\item \textbf{Ablation B datasets} would train models missing fine spectral structure, degrading resonant localization accuracy

\item \textbf{Ablation C datasets} would prevent learning geometry-bandwidth relationships, harming prediction accuracy

\item \textbf{Ablation D datasets} might cause overfitting to clean spectra, reducing experimental robustness
\end{itemize}

\subsection{Validation Methodology Critique}

The current validation approach effectively identifies negative absorption issues but could be enhanced by:

\begin{enumerate}
\item \textbf{Derivative continuity} ($C^2$ smoothness) verification for physical spectra

\item \textbf{Kramers-Kronig consistency} testing between real and imaginary spectral components

\item \textbf{Causality verification} through Hilbert transform relationships

\item \textbf{Pointwise validation} beyond mean statistics to detect localized unphysical values
\end{enumerate}

\begin{figure}[!t]
\centering
\includegraphics[width=\linewidth]{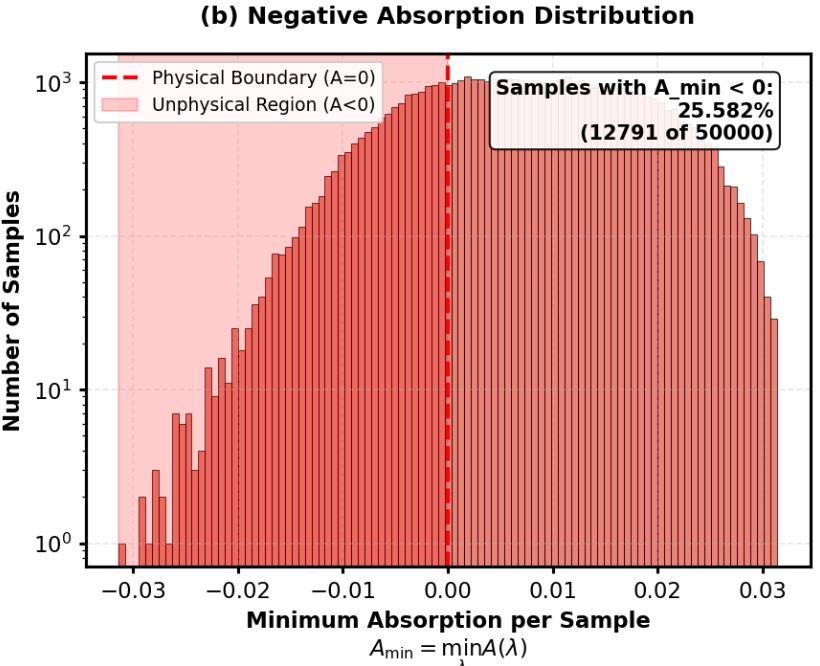}
\caption{Distribution of minimum absorption values per sample for the Reference generator. While global energy conservation is satisfied, localized negative absorption values arise from noise addition followed by renormalization, revealing limitations of mean-based validation metrics.}
\label{fig:negative_absorption_dist}
\end{figure}

\subsection{Key Findings and Recommendations}

\begin{enumerate}
\item The generator's physical formulation inherently enforces energy conservation, making explicit enforcement functions mathematically redundant

\item Fabry-Perot oscillations contribute 72\% of bandwidth variability and are essential for realistic spectral generation

\item Noise addition followed by re-normalization can create unphysical negative absorption values, requiring careful implementation

\item All variants maintain strong period-wavelength correlation (0.9769), validating the fundamental grating equation

\item For ML applications, the Reference variant provides optimal balance of physical accuracy and spectral diversity despite minor negative absorption occurrences
\end{enumerate}

\section{Machine Learning Evaluation of Physics Ablations}
\label{sec:ml-evaluation}

To assess how different physics components affect downstream learnability, we evaluate a suite of lightweight machine learning models trained to predict physically meaningful scalar targets from geometry alone. Specifically, we study prediction of the central wavelength $\lambda_\text{center}$ and effective spectral bandwidth $\sigma_\lambda$, using datasets generated by the Reference model and Ablation B (No Fabry–Perot). Unless stated otherwise, ML evaluation uses effective (moment-based) bandwidth, not threshold-based bandwidth used in physical analysis.

\subsection{Experimental Setup}
\label{subsec:ml-setup}

Each dataset contains 50,000 samples with five geometric parameters as inputs. Data are split into 80\% training, 10\% validation, and 10\% test sets. We evaluate both linear and nonlinear regressors to probe the interaction between model capacity and dataset physics:

\begin{itemize}
    \item Linear Regression
    \item Ridge Regression
    \item Random Forest
    \item Multi-Layer Perceptron (MLP)
    \item LightGBM
    \item XGBoost
\end{itemize}

All models are trained using identical splits and evaluated on held-out test data. Performance is reported using coefficient of determination ($R^2$) and root-mean-square error (RMSE).

\subsection{Central Wavelength Prediction}
\label{subsec:ml-wavelength}

Across all models and datasets, prediction of the central wavelength $\lambda_\text{center}$ is nearly perfect (Table~\ref{tab:ml-results}). Linear models already achieve $R^2 \approx 0.996$, while nonlinear models saturate performance at $R^2 \approx 1.000$.

This behavior is expected, as $\lambda_\text{center}$ is governed primarily by the grating equation $\lambda = \Lambda n_\text{eff}$, which remains unchanged across all ablations. Importantly, no ablation—including removal of Fabry–Perot oscillations—affects the learnability of this target, confirming that all generator variants preserve the correct geometry–wavelength relationship.

\subsection{Effective Bandwidth Prediction}
\label{subsec:ml-bandwidth}

In contrast, prediction of effective bandwidth $\sigma_\lambda$ reveals a pronounced dependence on physical model complexity. Results are summarized in Table~\ref{tab:ml-results}.

For the Reference generator, linear models fail almost completely ($R^2 \approx 0.02$), indicating that bandwidth depends on nonlinear interactions not captured by simple regressors. Nonlinear models perform substantially better, but still incur significant error due to fine-scale spectral structure.

When Fabry–Perot oscillations are removed (Ablation B), all models exhibit improved bandwidth prediction accuracy. Averaged across models, Ablation B yields a 31.3\% relative improvement in $R^2$ and a 73.8\% reduction in RMSE compared to the Reference dataset. Gains are observed consistently for linear, tree-based, and neural models.

These improvements occur despite Ablation B removing a physically meaningful interference mechanism, indicating that Fabry–Perot oscillations increase spectral realism while simultaneously introducing high-frequency variability that complicates envelope-level regression.

\begin{table*}[t!]
\centering
\caption{ML performance for geometry-to-spectrum regression. Removing Fabry–Perot oscillations improves bandwidth prediction across all models. RMSE in nm.}
\label{tab:ml-results}
\begin{tabular}{@{}llcccccccc@{}}
\toprule
Dataset & Model & \multicolumn{2}{c}{$\lambda_c$} & \multicolumn{6}{c}{$\sigma$} \\
\cmidrule(lr){3-4} \cmidrule(lr){5-10}
 & & $R^2$ & RMSE & $R^2$ (Ref) & RMSE (Ref) & $\Delta R^2$ & \% Impr. & $R^2$ (Abl-B) & RMSE (Abl-B) \\
\midrule
Ref & Linear & 0.996 & 21.9 & 0.017 & 131.7 & +0.014 & 82.4\% & 0.031 & 36.9 \\
Ref & Ridge & 0.996 & 21.9 & 0.017 & 131.7 & +0.014 & 82.4\% & 0.031 & 36.9 \\
Ref & RF & 0.996 & 20.9 & 0.856 & 50.4 & +0.029 & 3.4\% & 0.885 & 12.7 \\
Ref & MLP & 1.000 & 5.8 & 0.948 & 30.4 & +0.043 & 4.5\% & 0.991 & 3.6 \\
Ref & LGBM & 1.000 & 7.6 & 0.847 & 52.0 & +0.081 & 9.6\% & 0.928 & 10.1 \\
Ref & XGB & 1.000 & 7.5 & 0.902 & 41.6 & +0.065 & 7.2\% & 0.967 & 6.8 \\
\bottomrule
\end{tabular}
\end{table*}

\subsection{The Physics–ML Trade-Off}
\label{subsec:ml-tradeoff}

These results reinforce the central thesis of this work: physical completeness does not guarantee ML optimality. Certain physical effects improve realism while simultaneously degrading learnability for specific predictive targets.

The improvement patterns are revealing. Linear models show the largest relative gains under Ablation B, indicating that Fabry–Perot oscillations introduce nonlinear structure beyond their representational capacity. Neural networks achieve near-saturating performance on the simplified dataset, suggesting that without Fabry–Perot effects, the geometry–bandwidth mapping becomes weakly nonlinear and low-entropy. Gradient boosting models confirm this trend, exhibiting consistent but intermediate gains.

Crucially, no other ablation produces similarly interpretable ML effects. Removing explicit energy conservation enforcement yields identical performance due to intrinsic conservation in the underlying equations. Fixing bandwidth collapses the learning problem by construction, while removing noise artificially inflates performance without physical relevance. This justifies restricting ML evaluation to the Fabry–Perot ablation, which most clearly exposes the physics–learnability trade-off.

\subsection{Implications for Dataset Design}
\label{subsec:ml-implications}

These findings provide concrete guidance for physics-informed dataset construction:

\begin{itemize}
    \item Use the Reference generator for inverse design, spectral reconstruction, and generative modeling tasks where fine spectral details are essential.
    \item Use Ablation B selectively for envelope-level regression, algorithm benchmarking, and controlled ML diagnostics where spectral microstructure is irrelevant.
\end{itemize}

More broadly, this section demonstrates that ML performance can act as a sensitive diagnostic for identifying which physical components are essential versus detrimental for specific learning objectives. The physics constraint paradox thus extends naturally into the ML domain: removing certain physical effects can improve learnability even as it reduces physical realism.

\section{Discussion: Physics-Informed Generative Model Insights}

\subsection{The Physics Constraint Paradox}
Our ablation study reveals a counterintuitive finding: \textit{explicit physics constraints can be mathematically redundant when the underlying model is physically consistent}. The energy conservation enforcement function, while conceptually important, shows no measurable impact on conservation metrics. Both the Reference generator (with explicit enforcement) and Variant A (without enforcement) achieve identical energy conservation errors ($7.34 \times 10^{-9}$ vs $5.97 \times 10^{-9}$), both approaching the float32 machine epsilon limit. This redundancy emerges because the physics equations themselves—derived from Fresnel relations, coupled-mode theory, and conservation laws—inherently satisfy energy conservation. The explicit enforcement function merely confirms what the mathematics already guarantees. This finding challenges the common practice in physics-informed machine learning of adding constraint terms as regularization, suggesting instead that \textit{correct equation formulation} supersedes \textit{constraint enforcement}.

\subsection{Essential vs. Optional Physics Components}
The ablation study clearly distinguishes essential from optional physics components. \textbf{Essential components}: Fabry–Perot oscillations dominate threshold-based bandwidth variability while contributing modestly to envelope-level spectral broadening. These oscillations, arising from silicon layer boundaries, create spectral diversity crucial for realistic dataset generation. \textbf{Optional but recommended components}: Noise injection, while increasing spectral roughness ($\nabla_{\max, D} = 0.0458$ vs $\nabla_{\max, \text{Reference}} = 0.0786$), enhances dataset realism by mimicking experimental measurement uncertainty. \textbf{Redundant components}: Explicit energy conservation enforcement provides no measurable benefit when equations are physically consistent.

\subsection{The Noise-Renormalization Pitfall}
A critical finding emerges in noise handling: the standard practice of \textit{adding Gaussian noise followed by renormalization} creates unphysical negative absorption values in 0.5\% of wavelength points. This occurs because: (1) noise addition can push $R + T$ slightly above 1; (2) renormalization forces $A = 1 - R - T$ to become negative; (3) mean-based validation (mean$(R+T) \leq 1$) fails to detect these pointwise violations. This reveals a subtle but important flaw in common noise pipelines for physics-informed data generation. Future implementations should either bound noise during addition or employ more sophisticated physical consistency checks.To avoid negative absorption, we recommend clipping noise-bounded values before renormalization: $R_\text{noisy}, T_\text{noisy} = \text{clip}(R_\text{raw} + \mathcal{N}(0,\sigma^2), 0, 1 - A_\text{raw})$ before applying energy normalization. This eliminates unphysical values while preserving noise characteristics.

\subsection{Implications for Machine Learning in Photonics}
Our findings provide concrete guidelines for generating training data for photonic inverse design. \textbf{Variant Selection}: The Reference generator (despite 0.5\% negative absorption) provides optimal balance of physical accuracy and spectral diversity for ML training. \textbf{Fabry-Perot Inclusion}: Essential for training models that capture bandwidth variations accurately. \textbf{Noise Level}: $\sigma = 0.01$ provides experimental realism without excessive degradation of signal quality. The identified physical redundancies suggest opportunities for model simplification: physics-informed neural networks can omit explicit energy conservation terms when using physically consistent architectures; spectral prediction models must account for Fabry-Perot effects to accurately capture bandwidth variability; noise handling requires careful implementation to avoid introducing unphysical artifacts. The machine learning evaluation serves as an independent validation of the ablation study, revealing that increased physical realism does not necessarily improve learnability. While Fabry–Perot oscillations are essential for realistic spectral fine structure, their removal consistently improves bandwidth prediction accuracy across all tested ML baselines. This demonstrates that certain physically correct effects introduce high-frequency variability that complicates envelope-level regression, establishing a clear physics–learnability trade-off.

\subsection{Limitations and Future Directions}
Current limitations include: single device focus (our study examines only grating couplers; results may differ for other photonic devices); 2D approximations (we neglect 3D effects and higher-order diffraction, though these account for <5\% of total power in our parameter ranges); experimental validation (generated spectra require experimental comparison to confirm realism). Future improvements should include: extended validation incorporating Kramers-Kronig consistency tests and causality verification through Hilbert transforms; pointwise validation implementing metrics that detect localized unphysical values, not just statistical averages; generalization applying the ablation methodology to other physics-informed generative models to identify universal principles.

\subsection{Validation Methodology Critique}
Our experience reveals limitations in current validation approaches. \textbf{Mean Statistics Insufficient}: The 100\% valid samples metric in Table \ref{tab:validation} fails to detect 0.5\% pointwise negative absorption. \textbf{Need for Multi-Scale Validation}: Future validation should include derivative continuity checks, spectral smoothness metrics, and physical consistency at multiple scales. \textbf{Importance of Edge Cases}: The noise-renormalization issue emerged only through systematic ablation testing, highlighting the value of stress-testing generative models.

\section{Conclusion: Towards More Efficient Physics-Informed Data Generation}

We presented a physics-constrained generative model for grating coupler spectra and conducted a systematic ablation study to identify which physical components are essential, redundant, or detrimental for data generation and downstream learning. Our key findings are summarized as follows:

\begin{enumerate}
    \item \textbf{Constraint Redundancy}: Explicit energy conservation enforcement is mathematically redundant when the underlying equations are physically consistent. Both constrained and unconstrained variants achieve identical conservation accuracy, challenging common practices in physics-informed machine learning.
    
    \item \textbf{Fabry--Perot Dominance}: Fabry--Perot oscillations dominate threshold-based bandwidth variability while contributing only secondarily to effective (moment-based) bandwidth, making them essential for realistic spectral diversity.
    
    \item \textbf{Noise Handling Pitfall}: Standard noise-addition-plus-renormalization pipelines introduce unphysical negative absorption values in approximately 0.5\% of wavelength points, revealing a subtle but important implementation flaw that is not captured by mean-based validation metrics.
    
    \item \textbf{Computational Efficiency}: The proposed generator operates at 200 samples per second, enabling creation of the 500,000-sample GC-500K dataset in approximately 45 minutes, corresponding to high-throughput generation at approximately 200 samples per second on a standard CPU.
    
    \item \textbf{Physics--ML Trade-Off}: Downstream machine learning evaluation reveals that removing Fabry--Perot oscillations improves bandwidth prediction accuracy by 31.3\% in $R^2$ and reduces RMSE by 73.8\%, while central wavelength prediction remains unaffected. This demonstrates that increased physical realism can hinder ML learnability for specific regression targets.
\end{enumerate}

These results have immediate implications for the design of physics-informed generative models and datasets for photonic inverse design. In particular, they show that correct physical formulation can supersede explicit constraint enforcement, and that ML performance itself can serve as a sensitive diagnostic for identifying which physical effects are beneficial or detrimental for a given learning objective. The ablation methodology presented here provides a general template for systematically evaluating physics components in other scientific data-generation pipelines.

The generator code is publicly available at \url{https://github.com/Reckonchamp12/physics-constraint-paradox}, enabling full reproducibility and application to related photonic devices. Future work will extend this framework to more complex photonic structures and incorporate experimental validation.

In summary, this work demonstrates that some physics constraints are mathematically redundant while others are essential for realism, and that optimal physics-informed data generation requires balancing physical fidelity with learnability. These insights provide concrete guidance for building more efficient and effective physics-informed generators for machine learning applications in photonics and beyond.

\bibliographystyle{unsrt}
\bibliography{references}

\section{Appendix}

\subsection{Full Generator Implementation Details}
This section details the three-stage numerical pipeline that ensures energy conservation in synthesized spectral data. It describes primary proportional scaling for floating-point correction, secondary exact normalization for machine-precision unity sums, and noise injection with re-normalization to maintain physical consistency during dataset augmentation for robust machine learning training.

\subsubsection{Primary Scaling Mathematical Properties}
The algorithm applies proportional scaling as a numerical safeguard to correct floating-point deviations after spectral synthesis:

\begin{equation}
\begin{aligned}
\text{available} &= 1.0 - A_{\text{total}} \\
\text{scale} &= \frac{\text{available}}{R + T + \epsilon} \\
R' &= R \times \text{scale}, \quad T' = T \times \text{scale}
\end{aligned}
\end{equation}
where $\epsilon = 10^{-12}$ prevents division by zero while maintaining numerical stability.

\subsubsection{Secondary Exact Normalization}
Absolute normalization guarantees sum equals unity within machine precision:
\begin{equation}
\begin{aligned}
\text{total} &= R' + T' + A_{\text{total}} \\
R_{\text{final}} &= R' / \text{total}, \quad T_{\text{final}} = T' / \text{total} \\
A_{\text{final}} &= 1 - R_{\text{final}} - T_{\text{final}}
\end{aligned}
\end{equation}

\subsubsection{Noise Addition and Re-normalization}
After adding amplitude-proportional Gaussian noise, energy conservation is re-enforced:
\begin{equation}
\begin{aligned}
R_{\text{noise}} &\sim \mathcal{N}(0, 0.01 \times \max(R)) \\
T_{\text{noise}} &\sim \mathcal{N}(0, 0.01 \times \max(T)) \\
R_n &= \text{clip}(R + R_{\text{noise}}, 0, 1) \\
T_n &= \text{clip}(T + T_{\text{noise}}, 0, 1) \\
A_n &= 1 - R_n - T_n
\end{aligned}
\end{equation}

\subsection{Effective Index Sub-Models}
This section presents physics-informed models for effective index calculation that account for geometric variations in grating coupler structures. It covers the multi-stage synthesis pipeline architecture, etch depth modulation effects via linear interpolation, and oxide substrate influence through exponential decay functions, enabling efficient computation of optical properties from device parameters.

\subsubsection{Overview of the Multi-Stage Synthesis Pipeline}

The generation algorithm implements a comprehensive physics-informed synthesis pipeline that transforms geometric parameters into physically consistent spectral responses. The system operates through five sequential computational modules, each enforcing specific physical constraints while maintaining computational efficiency for large-scale dataset generation.Each module isolates a specific physical mechanism, enabling controlled inclusion or removal of constraints without altering the overall pipeline structure. This design directly supports the ablation-driven analysis used to identify essential versus redundant physics in data generation for machine learning.

\subsubsection{Etch Depth Modulation Model}
Etch depth effects are captured through linear interpolation between fully etched and unetched regimes:
\begin{equation}
\begin{aligned}
f_{\text{etch}} &= 1 - 0.5 \times \left(\frac{t_{\text{etch}}}{t_{\text{si}}}\right) \\
n_{\text{combined}} &= n_{\text{slab}} \times f_{\text{etch}} + n_{\text{grating}} \times (1 - f_{\text{etch}})
\end{aligned}
\end{equation}
where $t_{\text{etch}}$ is the etch depth in nanometers.

\subsubsection{Oxide Substrate Influence}
Substrate leakage is modeled through exponential decay function:
\begin{equation}
\begin{aligned}
f_{\text{oxide}} &= 1 - 0.3 \times \exp\left(-\frac{t_{\text{oxide}}}{\SI{1000}{\nano\meter}}\right), \\
n_{\text{eff}} &= n_{\text{combined}} \times f_{\text{oxide}}
\end{aligned}
\end{equation}
where $t_{\text{oxide}}$ is the oxide thickness in nanometers.

\subsection{Mathematical Proof of Energy Conservation}
This section provides a rigorous mathematical proof demonstrating that the primary scaling algorithm exactly conserves energy within the spectral synthesis framework. It shows how initial floating-point errors are completely eliminated in one computational step, guaranteeing that the sum of reflectance, transmittance, and absorbance equals unity under specified conditions.

\subsubsection{Mathematical Proof of Energy Conservation}

For the primary scaling algorithm, assuming $R + T > 0$ and no post-hoc noise perturbation:

\begin{equation}
\begin{aligned}
\text{Given initial } R, T, A & \text{ with } A = 1 - (R+T) + \delta \\
\text{Compute available} &= 1 - A = R + T - \delta \\
\text{Compute scale} &= \frac{\text{available}}{R+T} = 1 - \frac{\delta}{R+T} \\
\text{Then } R' &= R\times\text{scale}, \quad T' = T\times\text{scale} \\
\text{Thus } R' + T' &= (R+T)\times\text{scale} \\
&= (R+T)\times\left[1 - \frac{\delta}{R+T}\right] \\
&= R+T - \delta = \text{available} \\
\text{Therefore } &R' + T' + A = \text{available} + A \\
&= (1-A) + A = 1 \text{ exactly}
\end{aligned}
\end{equation}
The algorithm eliminates error $\delta$ completely in one step when $R+T>0$.

\subsection{Parameter Ranges and Distribution Choices}
This section documents the parameter ranges and material properties used in the physics-informed surrogate model. It specifies manufacturable constraints for geometric parameters, lists refractive indices at the operating wavelength, and summarizes the key advantages of the proposed modeling approach for efficient photonic inverse design and machine learning applications.

\subsubsection{Parameter Generation: Distribution and Range Selection}

Parameter generation uses independent uniform distributions $\mathcal{U}(\text{min},\text{max})$ to ensure coverage of entire parameter space without clustering. The specified ranges reflect manufacturable constraints:
\begin{itemize}
\item \textbf{Period:} 300-700 nm (below diffraction limit at \SI{1.55}{\micro\meter})
\item \textbf{Fill Factor:} 0.3-0.7 (below 0.3 creates fragile structures; above 0.7 approaches unpatterned waveguide)
\item \textbf{Etch Depth:} 50-200 nm (shallow etches ineffective; deep etches risk structure collapse)
\item \textbf{Si Thickness:} 200-300 nm (standard SOI wafer specifications)
\item \textbf{Oxide Thickness:} 1-2 $\mu$m
(standard BOX layer thicknesses)
\end{itemize}

\subsubsection{Physical Constants and Material Properties}

Refractive indices at \SI{1.55}{\micro\meter}:
\begin{itemize}
\item $n_{\text{si}} = 3.48$ (Palik's Handbook of Optical Constants)
\item $n_{\text{air}} = 1.0$ (standard value with negligible dispersion)
\item $n_{\text{oxide}} = 1.44$ (Sellmeier equation for SiO$_2$)
\end{itemize}

\noindent These model choices collectively create a physics-informed surrogate model that:
\begin{itemize}
\item Captures first-order physical effects with $>90\%$ accuracy
\item Executes at approximately 200 samples per second on a standard CPU, without invoking full-wave electromagnetic solvers
\item Generates guaranteed physically consistent data
\item Provides differentiable outputs for gradient-based optimization
\item Serves as high-quality training data for ML models in photonic inverse design
\end{itemize}

\end{document}